\documentclass[11pt,a4paper]{article}
\usepackage[hyperref]{naaclhlt2019}
\usepackage{times}
\usepackage{latexsym}
\usepackage{url}

\usepackage{xspace}
\usepackage{multirow}
\usepackage{hhline}
\usepackage{amsmath,amsfonts,amssymb}
\usepackage{pgfplots}
\pgfplotsset{compat=1.9}
\usepackage{graphicx}
\usepackage{xcolor}
\usepackage{float}
\restylefloat{table}
\usepackage[textsize=tiny]{todonotes}   %
\aclfinalcopy %
\def\aclpaperid{1346X-F3H4B3F3C5} %

\newcommand{\seqcube}{\textsc{seq}\textsuperscript{3}\xspace}
\newcommand{\rnn}{\textsc{rnn}\xspace}
\newcommand{\lstm}{\textsc{lstm}\xspace}

\newcommand{\seqseq}{\textsc{seq2seq}\xspace}
\newcommand{\oov}{\textsc{oov}\xspace}

\title{\textsc{SEQ}\textsuperscript{3}: Differentiable Sequence-to-Sequence-to-Sequence Autoencoder for Unsupervised Abstractive Sentence Compression}

\author{ 
	Christos Baziotis$^{1,2}$,  Ion Androutsopoulos$^2$, 
	Ioannis Konstas$^3$, Alexandros Potamianos$^1$ \\
	$^{1}$ School of ECE, National Technical University of Athens, Athens, Greece \\
	$^{2}$ Department of Informatics, Athens University of Economics and Business, Athens, Greece \\  
	$^{3}$ Interaction Lab, School of Math. and Comp. Sciences, Heriot-Watt University, Edinburgh, UK \\  
	{\tt cbaziotis@mail.ntua.gr, ion@aueb.gr} \\
	{\tt i.konstas@hw.ac.uk, potam@central.ntua.gr}
}

\date{}
\begin{document}
\maketitle

\begin{abstract}
Neural sequence-to-sequence models are currently the dominant approach in several natural language processing tasks, but require large parallel corpora. We present a sequence-to-sequence-to-sequence autoencoder (\textsc{seq}\textsuperscript{3}), consisting of two chained encoder-decoder pairs, 
with words used as a sequence of discrete latent variables.
We apply the proposed model to unsupervised abstractive sentence compression, where the first and last sequences are the input and reconstructed sentences, respectively, while the middle sequence is the compressed sentence. Constraining the length of the latent word sequences forces the model to distill important information from the input. A pretrained language model, acting as a prior over the latent sequences, encourages the compressed sentences to be human-readable. Continuous relaxations enable us 
to sample from categorical distributions, allowing gradient-based optimization, unlike alternatives that rely on reinforcement learning. The proposed model does not require parallel text-summary pairs, achieving 
promising results in unsupervised sentence compression 
on benchmark datasets.
\end{abstract}

\section{Introduction}
Neural sequence-to-sequence models (\seqseq) perform impressively well in several natural language processing tasks, such as machine translation~\cite{sutskever2014sequence, bahdanau2014neural} or syntactic constituency parsing~\cite{vinyals2015grammar}. However, they require massive parallel training datasets~\cite{koehn2017six}.
Consequently there has been extensive work on utilizing non-parallel corpora to boost the performance of \seqseq models~\cite{sennrich2016monolingual,gulcehre2015monolingual}, mostly in neural machine translation where models that require absolutely no parallel corpora have 
also been proposed~\cite{artetxe2018unsupervised, lample2018unsupervised}. 

Unsupervised (or semi-supervised) \seqseq models have also been proposed for summarization tasks with no (or small) parallel text-summary sets, including unsupervised sentence compression. Current models, however, barely reach lead-\textit{N} baselines~\cite{fevry2018, wang2018}, and/or are non-differentiable~\cite{wang2018, miao2016}, thus relying on reinforcement learning, which is unstable and inefficient.
By contrast, we propose a sequence-to-sequence-to-sequence autoencoder, dubbed \seqcube, that can be trained end-to-end via gradient-based optimization. \seqcube employs differentiable approximations for sampling from categorical distributions~\cite{maddison2017gumbel,jang2017categorical}, which have been shown to outperform reinforcement learning~\cite{havrylov2017}.
Therefore it is a generic framework which can be easily extended to other tasks, e.g., machine translation and semantic parsing via task-specific losses.  In this work, as a first step, we apply \seqcube to unsupervised abstractive sentence compression.

\begin{figure}[t]
	\includegraphics[clip, trim=0.2cm 1.2cm 0.2cm 1.9cm, width=1.0\columnwidth]{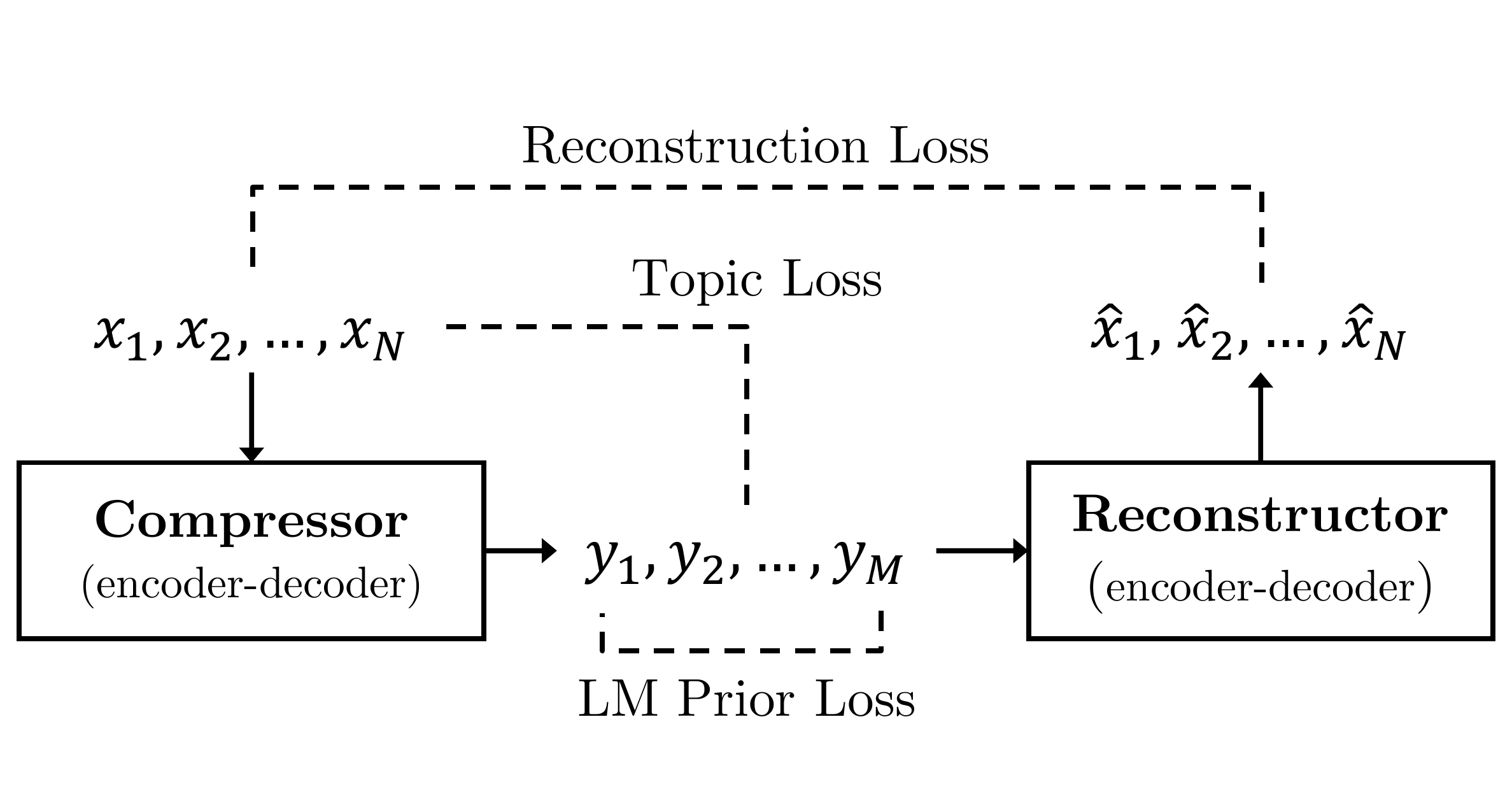}
	\vspace*{-7mm}
	\caption{Overview of the proposed \seqcube autoencoder.}
	\vspace*{-6mm}
	\label{fig:overview}
\end{figure}

\seqcube (\S\ref{sec:model}) comprises two attentional encoder-decoder~\cite{bahdanau2014neural} pairs (Fig.~\ref{fig:overview}): a compressor $C$ and a reconstructor $R$.  
$C$ (\S\ref{sec:compressor}) receives an input text $\mathbf{x}=\left<x_1,\ldots,x_N\right>$ of \textit{N} words, and generates a summary $\textbf{y}=\left<y_1,\ldots,y_M\right>$ of \textit{M} words (\textit{M}$<$\textit{N}), $\textbf{y}$ being a latent variable.  $R$ and $C$ communicate only through the discrete words of the summary $\textbf{y}$ (\S\ref{sec:diff_sampling}). $R$ (\S\ref{sec:reconstructor}) produces a sequence $\hat{\mathbf{x}}=\left<\hat{x}_1,\ldots,\hat{x}_N\right>$ of \textit{N} words from $\textbf{y}$, trying to minimize a reconstruction loss $L_{R}=(\mathbf{x},\hat{\mathbf{x}})$ (\S\ref{sec:losses}). A pretrained language model acts as a prior on $\textbf{y}$, introducing an additional loss $L_P(\textbf{x}, \textbf{y})$ that encourages \seqcube to produce human-readable summaries. A third loss $L_T(\textbf{x}, \textbf{y})$ rewards summaries $\textbf{y}$ with similar topic-indicating words as $\textbf{x}$. Experiments (\S\ref{sec:experiments}) on the Gigaword sentence compression dataset~\cite{rush2015giga} and the \textsc{duc-2003} and \textsc{duc-2004} shared tasks \cite{Over:2007:DC:1284916.1285157} 
produce promising results. 

Our contributions are: (1) a fully differentiable sequence-to-sequence-to-sequence (\seqcube) autoencoder that can be trained without parallel data via gradient optimization; (2) an application of \seqcube to unsupervised abstractive sentence compression, with additional task-specific loss functions; (3) state of the art performance in unsupervised abstractive sentence compression. This work is a step towards exploring the potential of \seqcube in other tasks, such as machine translation.%

\section{Proposed Model} \label{sec:model}

\begin{figure}[!t]
	\centering
	\includegraphics[clip, trim=7.95cm 1.0cm 7.95cm 1cm,
	width=1\columnwidth,page=4]{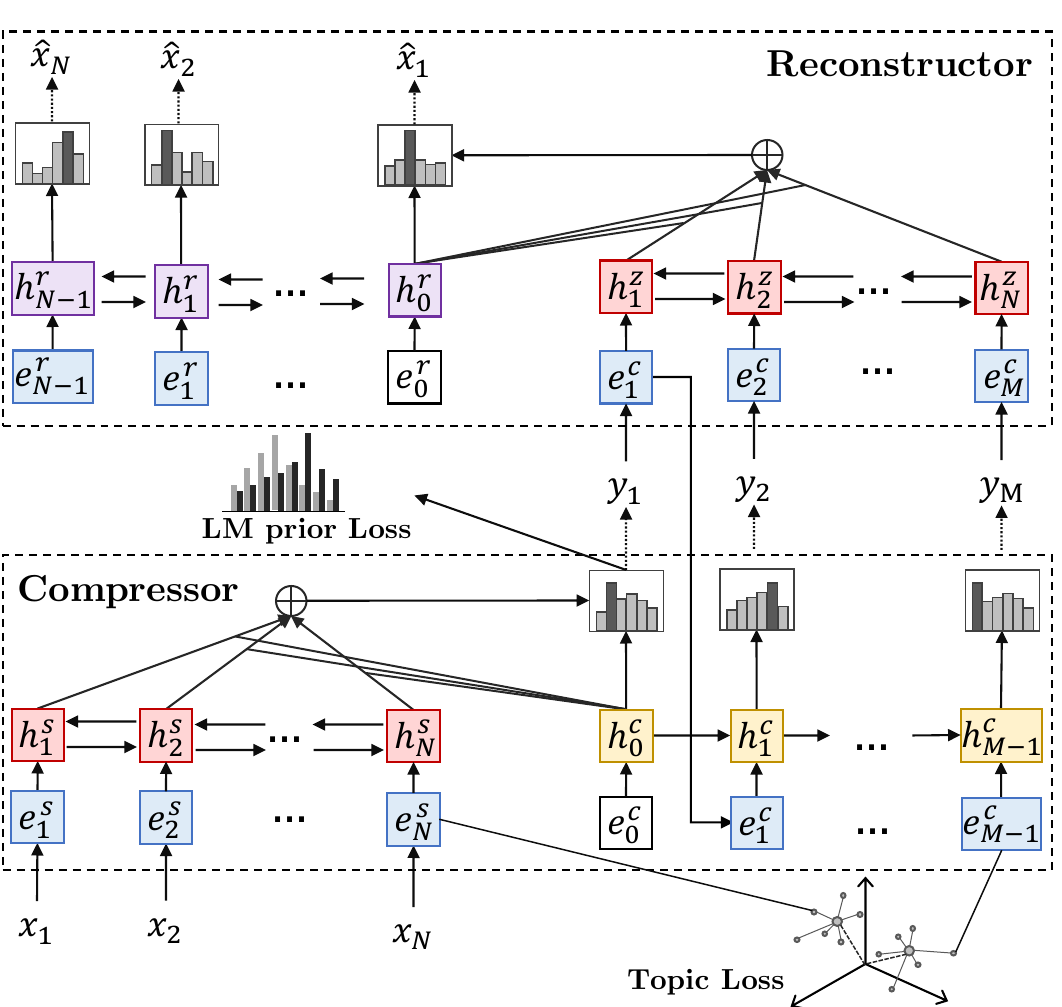}
	\vspace*{-6mm}
	\caption{More detailed illustration of \seqcube. The compressor ($C$) produces a summary from the input text, and the reconstructor ($R$) tries to reproduce the input from the summary. $R$ and $C$ comprise an attentional encoder-decoder each, and communicate only through the (discrete) words of the summary. The \textsc{lm} prior incentivizes $C$ to produce human-readable summaries, while topic loss rewards summaries with similar topic-indicating words as the input text. 
	}
	\vspace*{-4mm}
	\label{fig:architecture}
\end{figure}

\subsection{Compressor} \label{sec:compressor}

The bottom left part of Fig.~\ref{fig:architecture} illustrates the internals of the compressor $C$. An embedding layer projects the source sequence $\textbf{x}$ to the word embeddings $\textbf{e}^s=\left<e^s_1,\ldots,e^s_N\right>$, which are then encoded by a bidirectional \rnn, producing $\textbf{h}^s=\left<h^s_1,\ldots,h^s_N\right>$. Each $h^s_t$ is the concatenation of the corresponding left-to-right and right-to-left 
states (outputs in \lstm{s}) of the bi-\rnn.
\vspace*{-1ex}
\begin{align*}
    h^s_t &= 
    [\overrightarrow{\textsc{rnn}}_s (e^s_t, \overrightarrow{h}^s_{t-1});
    \overleftarrow{\textsc{rnn}}_s (e^s_t, \overleftarrow{h}^s_{t+1})]
\end{align*}

\noindent To generate the summary $\textbf{y}$, we employ the attentional \rnn decoder of \citet{luong2015}, with their global attention 
and input feeding. Concretely, at each timestep ($t \in \{1, \dots, \textit{M}\}$) we compute a probability distribution $a_i$ over all the states $h^s_1, \dots, h^s_N$ of the source encoder conditioned on the current state $h^c_t$ of the compressor's decoder to produce a context vector $c_t$.
\vspace*{-1.5ex}
    \[ a_i = 
    \textrm{softmax}({h^s_i}^\intercal \, W_a \, h^c_t), \;\;
    c_t = \sum_{i=1}^{N} a_i \, h^s_i \]
The matrix $W_a$ is learned. 
We obtain a probability distribution for $y_t$ over the vocabulary $V$ by combining $c_t$ and the current state $h^c_t$ of the decoder.
\begin{align}
    o^c_t &= \textrm{tanh}(W_o \, [c_t; h^c_t] + b_o) \label{eq:oct}\\ 
    u^c_t &= W_v \, o^c_t + b_v \label{eq:logit}\\ 
    p(y_t|y_{<t},\textbf{x}) &= \textrm{softmax}(u^c_t)\label{eq:pyt}
\end{align}
$W_o, b_o, W_v, b_v$ are learned. $c_t$ is also used when updating the state $h^c_t$ of the decoder, along with the embedding $e^c_t$ of $y_t$ and a countdown argument $\textit{M} - t$ (scaled by a learnable $w_d$) indicating the number of the remaining words of the summary \cite{fevry2018,kikuchi2016controlling}.
\begin{equation}
    h^c_{t+1} = 
    \overrightarrow{\textsc{rnn}}_c(h^c_t, e^c_t, c_t, w_d \, (\textit{M} - t))
    \label{eq:hc_update}
\end{equation}
For each input $\mathbf{x} = \left<x_1, \dots, x_N\right>$, 
we obtain a target length \textit{M} for the summary $\mathbf{y} = \left<y_1, \dots, y_M\right>$ by sampling (and rounding) from a uniform distribution $U(\alpha \textit{N}, \beta \textit{N})$; $\alpha, \beta$ are hyper-parameters ($\alpha < \beta < 1$); we set $\textit{M} = 5$, if the sampled \textit{M} is smaller. Sampling \textit{M}, instead of using a static compression ratio, allows us to train a model capable of producing summaries with varying (e.g., user-specified) compression ratios. Controlling the output length in encoder-decoder architectures has been explored in machine translation \cite{kikuchi2016controlling} and summarization \cite{fan2017length}.

\subsection{Differentiable Word Sampling} \label{sec:diff_sampling}
To generate the summary, we need to sample its words $y_t$ from the categorical distributions $p(y_t|y_{<t},\textbf{x})$, which is a  non-differentiable process. 

\noindent\textbf{Soft-Argmax}
Instead of sampling
$y_t$, a simple workaround 
during training is to pass as input to the next timestep 
of $C$'s decoder and to the corresponding timestep of $R$'s encoder
a weighted sum of all the vocabulary's ($V$) word embeddings, using a peaked softmax function~\cite{goyal2017credit}:
\begin{equation}
    \overline{e}^c_t = \sum_{i}^{|V|}\, e(w_i)\, \textrm{softmax}(u_t^c / \tau) 
    \label{eq:softargmax}
\end{equation}
where $u_t^c$ is the unnormalized score in Eq.~\ref{eq:logit} (i.e.,
the logit) of each word $w_i$ and $\tau\in(0,\infty)$ is the temperature.
As $\tau\rightarrow 0$ most of the probability mass in Eq.~\ref{eq:softargmax} goes to the most probable word, hence the operation approaches the $\arg\max$.

\smallskip 
\noindent\textbf{Gumbel-Softmax} 
We still want to be able to perform sampling, though, 
as it has the benefit of adding stochasticity and facilitating exploration of the parameter space.
Hence, we use the Gumbel-Softmax (\textsc{gs}) reparametrization trick~\cite{maddison2017gumbel,jang2017categorical} as a low variance approximation of sampling from categorical distributions. 
Sampling a specific word $y_t$ from the softmax (Eq.~\ref{eq:pyt}) is
equivalent to adding (element-wise) to the logits an independent noise sample $\xi$ from the Gumbel distribution\footnote{$\xi_i = -\log(-\log( x_i)), \;\; x_i \sim U(0,1) \nonumber $} and taking the $\arg\max$:
\begin{equation}
 y_t \sim \textrm{softmax}(u_t^c) 
 \leftrightarrow y_t = \arg\max(u_t^c+ \xi) 
 \label{eq:argmaxgs}
\end{equation}
Therefore, using the \textsc{gs} trick, Eq.~\ref{eq:softargmax} becomes:
\begin{equation}
    \tilde{e}^c_t = \sum_{i}^{|V|}\, e(w_i)\, \textrm{softmax}((u_t^c+ \xi) / \tau) 
    \label{eq:gs-sampling}
\end{equation}

\noindent\textbf{Straight-Through} Both relaxations lead to 
mixtures of embeddings, which do not correspond to actual words. Even though this enables the compressor to communicate with the reconstructor using continuous values, thus fully utilizing the available embedding space, ultimately our aim is to constrain them to communicate using \textit{only} natural language. 
In addition, an unwanted discrepancy is created between training (continuous embeddings) and test time (discrete embeddings). 
We alleviate these problems with the Straight-Through estimator 
(\textsc{st}) \cite{bengio2013st}. Specifically, in the forward pass of 
training we discretize $\tilde{e}^c_t$ by 
using the $\arg\max$ (Eq.~\ref{eq:argmaxgs}), whereas in the backward pass we compute the gradients using the 
\textsc{gs} (Eq.~\ref{eq:gs-sampling}). This is a biased estimator due to the mismatch between the forward and 
backward passes, but works well in practice.
\textsc{st gs} reportedly outperforms scheduled sampling~\cite{goyal2017credit} and converges faster than reinforcement learning~\cite{havrylov2017}.

\subsection{Reconstructor} \label{sec:reconstructor}
The reconstructor (upper right of Fig.~\ref{fig:architecture}) works like the compressor, but its encoder operates on the embeddings $e^c_1, \dots, e^c_M$ of the words $y_1, \dots, y_M$ of the summary (exact embeddings of the sampled words $y_t$ in the forward pass, approximate differentiable embeddings in the backward pass).

\subsection{Decoder Initialization} 
We initialize the hidden state of each decoder using a transformation of the concatenation $[\overrightarrow{h^s_N};\overleftarrow{h^s_1}]$ of the last hidden states (from the two directions) of its bidirectional encoder and a length vector, following \citet{mallinson2018length}. The length vector for the decoder of the compressor $C$ consists of the target summary length \textit{M}, scaled by a learnable parameter $w_v$, and the compression ratio $\frac{M}{N}$. 
\begin{align*}
    h^c_0 &= 
    \textrm{tanh} ( W_c \, 
        [ \overrightarrow{h^s_N}; \overleftarrow{h^s_1} ; 
          w_v M ; \dfrac{M}{N}])
\end{align*}
$W_c$ is a trainable hidden layer. The decoder of the reconstructor $R$ is initialized similarly.

\subsection{Loss Functions} \label{sec:losses}

\noindent\textbf{Reconstruction Loss}
$L_R(\textbf{x}, \hat{\textbf{x}})$ is the (negative) log-likelihood assigned by the (decoder of) 
$R$ to the input (correctly reconstructed) words $\textbf{x} = \left<x_1, \dots, x_N\right>$, where $p_R$ is the distribution of $R$. 
\[
    L_R(\textbf{x}, \hat{\textbf{x}}) = 
        - \sum_{i=1}^{N} 
        \log p_R(\hat{x}_i = x_i) 
\]
We do not expect 
$L_R(\textbf{x}, \hat{\textbf{x}})$ to decrease to zero, as there is information loss through the compression. However, we expect 
it to drive the compressor to produce such sentences that will increase the likelihood of the target words in the reconstruction.

\smallskip
\noindent\textbf{LM Prior Loss}
To ensure that the summaries $\textbf{y}$ are readable, we pretrain an \rnn language model (see Appendix) on the \textit{source} texts of the full training set. We compute the Kullback-Leibler divergence $D_\textrm{KL}$ between the probability distributions of the (decoder of) the compressor ($p(y_t|y_{<t},\textbf{x})$,  Eq.~\ref{eq:pyt}) and the language model ($p_{\textrm{LM}}(y_t|y_{<t})$).
Similar priors have been used in sentence compression~\cite{miao2016} and agent communication~\cite{havrylov2017}.

\smallskip
\noindent We also use the following task-specific losses.

\smallskip
\noindent\textbf{Topic Loss}
Words with high \textsc{tf-idf} scores are indicative of the topic of a text~\cite{ramos2003using,erkan2004lexrank}. 
To encourage the compressor to preserve in the summary $\textbf{y}$ the topic-indicating words of the input $\textbf{x}$, we compute the \textsc{tf-idf}-weighted average $v^x$ of the word embeddings of $\textbf{x}$ and
the average $v^y$ of the word embeddings of $\textbf{y}$ and use their cosine distance as an additional loss $L_T = 1 - 
\cos(v^x, v^y)$.
\begin{align*}
    v^x = \sum_{i=1}^N
    \dfrac{\textrm{\textsc{idf}}(x_i) \, e^s_i}{\sum_{t=1}^{N} \textrm{\textsc{idf}}(x_t)} \;\;\;
    v^y = \frac{1}{M} \sum_{i=1}^M e^c_i 
\end{align*}
(Using \textsc{tf-idf} in $v^y$ did not help.) 
All \textsc{idf} scores are computed on the training set. 

\smallskip
\noindent\textbf{Length Penalty} A fourth loss $L_L$ (not shown in Fig.~\ref{fig:overview}) helps the (decoder of the) compressor 
to predict the end-of-sequence (\textsc{eos}) token at the target summary length \textit{M}. $L_L$ is the cross-entropy between the distributions $p(y_t|y_{<t},\textbf{x})$ (Eq.~\ref{eq:pyt}) of the compressor at $t = \textit{M}+1$ and onward, with the one-hot distribution of the \textsc{eos} token.

\subsection{Modeling Details}
\noindent\textbf{Parameter Sharing} 
We tie the weights of layers encoding similar information, 
to reduce the number of trainable parameters. 
First, we use a shared embedding layer for 
the encoders and decoders, 
initialized with 100-dimensional GloVe embeddings~\cite{pennington2014glove}. 
Additionally, we tie the shared embedding layer with the output layers of both decoders~\cite{E17-2025, Inan2017TyingWV}. 
Finally, we tie the encoders of the compressor 
and reconstructor (see Appendix).

\smallskip
\noindent\textbf{OOVs} Out-of-vocabulary words are handled %
as in \citet{fevry2018} (see Appendix).
\begingroup
\captionsetup[figure]{font=small,skip=0pt}
\setlength{\tabcolsep}{12pt} %
\renewcommand{\arraystretch}{1.0} %
\begin{table*}[!h]
\centering
\small
\begin{tabular}{@{~}l@{~}@{~}c@{~}@{~}l@{~}ccc}
\hline
\textbf{Type} & \textbf{Supervision} & \textbf{Methods} & \textbf{R-1} & \textbf{R-2} & \textbf{R-L} \\\hline
\multirow{3}{*}{Supervised} & \multirow{3}{*}{3.8M} & ABS~\cite{rush2015giga} & 29.55 & 11.32 & 26.42 \\
 &  & SEASS~\cite{zhou2017selective}& 36.15 & 17.54 & 33.63 \\
 &  & words-lvt5k-1sent~\cite{nallapati2016abstractive} & \textbf{36.4} & \textbf{17.7} & \textbf{33.71} \\ 
 \hline
\multirow{1}{*}{Weakly supervised} & \multirow{1}{*}{(3.8M)} & Adversarial REINFORCE~\cite{wang2018}\hspace{10pt}  & 28.11 & 9.97 & 25.41 \\ \hline
\multirow{6}{*}{Unsupervised} & \multirow{6}{*}{0} 
& \textsc{Lead-8} (Baseline) & 21.86 & 7.66 & 20.45 \\
& &   Pretrained Generator~\cite{wang2018} & 21.26 & 5.60 & 18.89 \\
&  &   \seqcube (Full)                           & \textbf{25.39} & \textbf{8.21}    & \textbf{22.68} \\
&  &   \seqcube w/o \textsc{lm} prior loss       & 24.48          & 6.68             & 21.79 \\
&  &   \seqcube w/o \textsc{topic} loss    & 3.89           & 0.1              & 3.75 \\
\hline
\end{tabular}
\vspace*{-2mm}
\caption{Average results on the (English) Gigaword dataset for abstractive sentence compression methods.}
\label{table:results}
\end{table*}
\endgroup
\begingroup
\setlength{\tabcolsep}{8pt} %
\renewcommand{\arraystretch}{1.1} %
\begin{table}[tbh]
\centering
\small
\begin{tabular}{@{~}l@{~}ccc} \hline
\textbf{Model}     & \textbf{R-1}   & \textbf{R-2}  & \textbf{R-L}   \\ \hline
\textsc{abs}~\cite{rush2015giga}  \hspace{20pt} &  28.48 & 8.91 & 23.97 \\\hline 
\textsc{Prefix}    & \textbf{21.3} & \textbf{6.38} & \textbf{18.82} \\
\seqcube (Full) & 20.90 & 6.08 & 18.55 \\ 
\hline 
\end{tabular}
\vspace*{-5pt}
\caption{Averaged results on the \textsc{duc}-2003 dataset; the top part reports results of supervised systems.}
\label{table:duc2003}
\end{table}
\endgroup
\begingroup
\setlength{\tabcolsep}{7pt} %
\renewcommand{\arraystretch}{1.1} %
\begin{table}[tbh]
\centering
\small
\footnotesize
\begin{tabular}{@{~}l@{~}@{~~}ccc@{~~}} \hline
\textbf{Model}     & \textbf{R-1}   & \textbf{R-2}  & \textbf{R-L}   \\ \hline
\textsc{Topiary}~\cite{Zajic:2007:MRS:1284916.1285161} \hspace{3pt} & 25.12 & 6.46 & 20.12 \\
\citet{D10-1050} & 22 & 6 & 17 \\
\textsc{abs}~\cite{rush2015giga}     &  28.18 & 8.49 & 23.81 \\\hline 
\textsc{Prefix}     & 20.91 & 5.52 & 18.20 \\
\seqcube (Full) & \textbf{22.13} & \textbf{6.18} & \textbf{19.3} \\ 

\hline
\end{tabular}
\vspace*{-5pt}
\caption{Averaged results on the \textsc{duc}-2004 dataset; the top part reports results of supervised systems.}
\label{table:duc2004}
\end{table}
\endgroup

\section{Experiments} \label{sec:experiments}

\noindent\textbf{Datasets}
We train \seqcube on the \textit{Gigaword} sentence compression dataset~\cite{rush2015giga}.\footnote{\url{github.com/harvardnlp/sent-summary}} It consists of pairs, each containing the first sentence of a news article ($\textbf{x}$) and the article's headline ($\textbf{y}$), a total of 3.8M/189k/1951 train/dev/test pairs. 
We also test (without retraining) \seqcube on \textsc{duc}-2003 and \textsc{duc}-2004 shared tasks \cite{Over:2007:DC:1284916.1285157}, containing 624/500 news articles each, paired with 4 reference summaries capped at 75 bytes.

\smallskip
\noindent\textbf{Methods compared} 
We evaluated \seqcube and an ablated version of \seqcube.
We only used the article sentences (sources) of the training pairs from Gigaword to train \seqcube; 
our model is \textit{never} exposed to target headlines (summaries) 
during training or evaluation, i.e., it
is completely unsupervised. 
Our code is publicly available.\footnote{\url{https://github.com/cbaziotis/seq3}}
We compare \seqcube to other unsupervised sentence compression models. 
We note that the extractive model of \citet{miao2016} relies on a pre-trained attention model using at least 500K parallel sentences, which is crucial to mitigate the inefficiency 
of sampling-based variational inference and 
\textsc{reinforce}. Therefore it is not comparable, as it is semi-supervised.
The results of the extractive model of \citet{fevry2018} are also not comparable, as they were obtained on a different, not publicly available test set.
We note, however, that they report that their system performs worse than the \textsc{Lead-8} baseline in \textsc{rouge-2} and \textsc{rouge-l} on Gigaword. 
The only directly comparable unsupervised model is the abstractive `Pretrained Generator' of \citet{wang2018}. The version of `Adversarial \textsc{Reinforce}' that \citet{wang2018} consider unsupervised is actually weakly supervised, since its discriminator was exposed to the summaries of the same sources the rest of the model was trained on.

As baselines, we use \textsc{lead-8} for Gigaword, which simply selects the first 8 words of the source, and \textsc{prefix} for \textsc{duc}, which includes the first 75 bytes of the source article.
We also compare to supervised abstractive sentence compression methods (Tables~\ref{table:results}-\ref{table:duc2004}). Following previous work, we report the average F1 of \textsc{rouge-1}, \textsc{rouge-2}, \textsc{rouge-l} \cite{lin2004rouge}. 
We implemented \seqcube with \lstm{s}
(see Appendix) and during inference we perform greedy-sampling. 

\smallskip
\noindent\textbf{Results} Table~\ref{table:results} reports the Gigaword results.
\seqcube outperforms the unsupervised Pretrained Generator across all metrics by a large margin. It also surpasses \textsc{lead-8}.
If we remove the \textsc{lm} prior,  
performance drops, esp.\ in \textsc{rouge-2} and \textsc{rouge-l}.
This makes sense, since the pretrained \textsc{lm} 
rewards correct word order.
We also tried removing the topic loss, but the model failed to converge 
and results were extremely poor (Table~\ref{table:results}).
Topic loss acts as a bootstrap mechanism, biasing the compressor 
to generate words that maintain the topic of the input text. This greatly reduces variance due to sampling in early stages of training, alleviating the need to pretrain individual components, unlike works that rely on reinforcement learning~\cite{miao2016,wang2018}. 
Overall, both losses work in synergy, with the topic loss driving \textit{what} and the \textsc{lm} prior loss driving \textit{how} 
words should be included in the summary.
\seqcube behaves similarly on \textsc{duc}-2003 and \textsc{duc}-2004 (Tables~\ref{table:duc2003}-\ref{table:duc2004}), although it was trained on Gigaword.
In \textsc{duc}-2003, however, it does not surpass the \textsc{prefix} baseline.

Finally, Fig.~\ref{fig:examples} illustrates 
three randomly sampled outputs of \seqcube on Gigaword. In the first one, \seqcube copies several words esp.\ from the beginning of the input (hence the high \textsc{rouge-l}) exhibiting extractive capabilities, though still being adequately abstractive (bold words denote paraphrases). In the second one, \seqcube showcases its true abstractive power by paraphrasing and compressing multi-word expressions to single content words more heavily, still without losing the overall meaning. In the last example, \seqcube progressively becomes ungrammatical though interestingly retaining some content words from the input.
\begin{figure}[tb]
\footnotesize

\newcommand{\samplesize}{0.95}

\fbox{\begin{minipage}{\samplesize\columnwidth}
\textbf{input:} the american sailors who \textbf{thwarted} somali pirates flew home to the u.s. on wednesday but without their captain , who was still aboard a navy destroyer after being rescued from the \textbf{hijackers} .
\\
\textbf{gold}: us sailors who thwarted pirate hijackers fly home
\\
\textbf{\textsc{seq}\textsuperscript{3}}: the american sailors who \textbf{foiled} somali pirates flew home after crew \textbf{hijacked} .
\end{minipage}}

\fbox{\begin{minipage}{\samplesize\columnwidth}
\textbf{input:} the central election commission -lrb- cec -rrb- on monday \textbf{decided that taiwan will hold another election} of national assembly members on may \# .
\\
\textbf{gold}: national \texttt{<unk>} election scheduled for may
\\
\textbf{\textsc{seq}\textsuperscript{3}}: the central election commission -lrb- cec UNK \textbf{announced elections} .
\end{minipage}}

\fbox{\begin{minipage}{\samplesize\columnwidth}
\textbf{input:} dave bassett resigned as manager of struggling english premier league side nottingham forest on saturday after they were knocked out of the f.a. cup in the third round , according to local reports on saturday .
\\
\textbf{gold}: forest manager bassett quits .
\\
\textbf{\textsc{seq}\textsuperscript{3}}: dave bassett resigned as manager of struggling english premier league side UNK forest on knocked round press
\end{minipage}}
\vspace*{-5pt}
\caption{Good/bad example summaries on Gigaword.}
\label{fig:examples}

\end{figure}
\section{Limitations and Future Work} \label{sec:conclusions}
The model tends to copy the first words of the input sentence in the
compressed text (Fig.~\ref{fig:examples}).
We hypothesize that since the reconstructor 
is autoregressive, i.e., each word is conditioned on the previous one, errors occurring early in the generated sequence have cascading effects.
This inevitably encourages the compressor to select the first words of the input.
A possible workaround might be to 
modify \seqcube so that the 
first encoder-decoder pair would turn the inputs to longer sequences, and the second encoder-decoder would compress them trying to reconstruct the original inputs.
In future work, we plan to explore the potential of \seqcube in other tasks, such as unsupervised machine translation~\cite{lample2018word,artetxe2018unsupervised} and caption generation~\cite{pmlr-v37-xuc15}.
\section*{Acknowledgments}
We would like to thank Ryan McDonald for helpful discussions and feedback.
This work has been partially supported by computational time granted from the Greek Research \& Technology Network
(\textsc{GR-NET}) in the National \textsc{HPC} facility - \textsc{ARIS}. We thank  \textsc{NVIDIA} for donating a TitanX \textsc{GPU}.

\bibliography{refs}
\bibliographystyle{acl_natbib}

\newpage
\appendix
\section{Appendix}

\subsection{Temperature for Gumbel-Softmax}
Even though the value of the temperature $\tau$ does not affect the forward pass, it greatly affects the gradient computation and therefore the learning process. 
\citet{jang2017categorical} propose to anneal $\tau$ during training towards zero.
\citet{gulcehre2017memory} propose to learn $\tau$ as a function of the compressor's decoder state $h^c_t$, in order to reduce hyper-parameter tuning:
\begin{equation}
 \tau(h^c_t) =  \frac{1}{\log(1 + \exp(w_{\tau}^\intercal \, h^c_t)) + 1}
 \label{eq:tau}
\end{equation}
where $w_{\tau}$ is a trainable parameter and $\tau(h^c_t)\in(0,1)$.
\citet{havrylov2017} add $\tau_0$ as a hyper-parameter which controls the upper bound of the temperature. 
\begin{equation}
 \tau(h^c_t) =  \frac{1}{\log(1 + \exp(w_{\tau}^\intercal \, h^c_t)) + \tau_0}
 \label{eq:tau0}
\end{equation}
\begin {figure}[h]
    \centering
    \resizebox {\columnwidth} {!} {
            \begin{tikzpicture}
                \begin{axis}[
                    axis lines = center,
                    ytick={0,0.5, 1, ...,2},
                    ymax=2.1,
                ]

                \addplot [
                    thick,
                    domain=-8:8, 
                    samples=100, 
                    color=green,
                    ]
                {1 / (ln(1 + e^x) + 0.5) };
                \addlegendentry{$\tau_0=  0.5$}
                
                \addplot [
                    thick,
                    domain=-8:8, 
                    samples=100, 
                    color=red,
                ]
                {1 / (ln(1 + e^x) + 1) };
                \addlegendentry{$\tau_0= 1\,\,\,\,$}
                
                \addplot [
                    thick,
                    domain=-8:8, 
                    samples=100, 
                    color=blue,
                    ]
                {1 / (ln(1 + e^x) + 2) };
                \addlegendentry{$\tau_0=2\,\,\,\,$}
                 
                \end{axis}
        \end{tikzpicture}
    }
\caption{Plot of Eq.~\ref{eq:tau0}, with different values for the upper bound $\tau_0$.}
\end {figure}
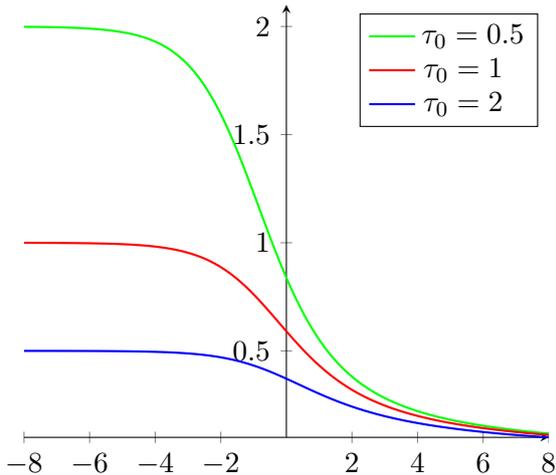

In our experiments, we had convergence problems with the learned temperature technique. We found that the compressor preferred values close to the upper bound, which led to unstable training, forcing us to set $\tau_0>1$ to stabilize the training process. Our findings align with the behavior reported by \citet{DBLP:conf/aaai/GuIL18}. Consequently, we follow their choice and fix $\tau=0.5$, which worked well in practice.

\subsection{Out of Vocabulary (OOV) Words}
The vocabulary of our experiments comprises the 15k most frequent words of Gigaword's training input texts (without looking at their summaries). To handle \oov{s}, we adopt the approach of \citet{fevry2018}, which can be thought of as a simpler form of copying compared to pointer networks \cite{see2017pointer}. We use a small set (10 in our experiments) of special \oov tokens $\oov_1$, $\oov_2$, \dots, $\oov_{10}$, whose embeddings are updated during learning. Given an input text $\textbf{x} = \left<x_1, \dots, x_N \right>$, we  replace (before feeding $\textbf{x}$ to \seqcube) each unknown word $x_i$ with the first unused (for the particular $\textbf{x}$) \oov token, taking care to use the same \oov token for all the occurrences of the same unknown word in $\textbf{x}$. For example, if `John' and `Rome' are not in the vocabulary, then ``John arrived in Rome yesterday. While in Rome, John had fun.''  becomes ``$\oov_1$ arrived in $\oov_2$ yesterday. While in $\oov_2$, $\oov_1$ had fun.'' If a new unknown word $x_i$ is encountered in $\textbf{x}$ and all the available \oov tokens have been used, $x_i$ is replaced by `\textsc{unk}', whose embedding is also updated during learning. The \oov tokens (and `\textsc{unk}') are included in the vocabulary, and \seqcube learns to predict them as summary words, in effect copying the corresponding unknown words of $\textbf{x}$. 
At test time, we replace the \oov tokens with the corresponding unknown words.

\subsection{Reconstruction Word Drop}\label{sec:word-drop}
Our model is an instance of Variational Auto-Encoders (\textsc{vae}) \cite{DBLP:journals/corr/KingmaW13}. A common problem in \textsc{vae}s is that the reconstructor tends to disregard the latent variable. 
We weaken the reconstructor $R$, in order to force it to fully utilize the latent sequence $\mathbf{y}$ to generate $\mathbf{\hat{x}}$.
To this end, we employ word dropout as in \citet{K16-1002} and randomly drop a percentage of the input words, thus forcing $R$ to rely solely on $\mathbf{y}$ to make good reconstructions.

\subsection{Implementation and Hyper-parameters}

We implemented \seqcube in PyTorch~\cite{paszke2017automatic}. All the \rnn{s} are
\lstm{s}~\cite{hochreiter1997lstm}. We use a shared encoder for the compressor and the reconstructor, consisting of a two-layer bidirectional \lstm with size $300$ per direction. We use separate decoders for the compressor and the reconstructor; each decoder is a two-layer unidirectional \lstm with size $300$.
The (shared) embedding layer of the compressor and the reconstructor is initialized with $100$-dimensional GloVe embeddings~\cite{pennington2014glove} and is tied with the output (projection) layers of the decoders and jointly finetuned during training. 
We apply layer normalization~\cite{ba2016layer} to the context vectors (Eq.~\ref{eq:oct}) of the compressor and the reconstructor.
We apply word dropout~(\S\ref{sec:word-drop}) to the reconstructor with $p=0.5$.

During training, the summary length \textit{M} is sampled from $U(0.4\,\textit{N}, 0.6\,\textit{N})$; during testing, $\textit{M}=0.5 \, \textit{N}$. 
The four losses are summed,  
$\lambda$s being scalar hyper-parameters.
\begin{align*}
    L = \lambda_R\, L_R + \lambda_P \, L_P + \lambda_T \, L_T + \lambda_L L_L
\end{align*}
We set $\lambda_R = \lambda_T = 1$, $\lambda_L = \lambda_P = 0.1$. We use the Adam~\cite{kingma2014Adam} optimizer, with batch size 128 and the default learning rate $0.001$. The network is trained for 5 epochs.

\smallskip
\noindent\textbf{LM Prior} The pretrained language model is a two-layer \lstm of size $1024$ per layer. It uses its own embedding layer of size $256$, which is randomly initialized and updated when training the language model. We apply dropout with $p=0.2$ to the embedding layer and dropout with $p=0.5$ to the \lstm layers. We use Adam~\cite{kingma2014Adam} with batch size 128 and the network is trained for 30 epochs. The learning rate is set initially to $0.001$ and is multiplied with $\gamma=0.5$ every 10 epochs.

\smallskip
\noindent\textbf{Evaluation} Following \citet{chopra2016comp}, we filter out pairs with empty headlines from the test set.
We employ the \textsc{pyrouge} package with ``\textit{-m -n 2 -w 1.2}'' to compute \textsc{rouge} scores. We use the provided tokenizations of the Gigaword and \textsc{duc}-2003, \textsc{duc}-2004 datasets. All hyper-parameters were tuned on the development set.

\end{document}